\documentclass{article} 
\usepackage{iclr2026_conference,times}

\usepackage{amsmath,amsfonts,bm}

\def\eqref#1{equation~\ref{#1}}

\def\1{\bm{1}}

\DeclareMathAlphabet{\mathsfit}{\encodingdefault}{\sfdefault}{m}{sl}
\SetMathAlphabet{\mathsfit}{bold}{\encodingdefault}{\sfdefault}{bx}{n}

\usepackage{multirow}
\usepackage{hyperref}
\usepackage{url}
\usepackage{graphicx}
\usepackage{booktabs} 
\usepackage{xcolor}
\usepackage{amssymb}

\title{Retrieval Mechanisms Surpass Long-Context Scaling in Time Series Forecasting}

\author{Rishi Ahuja$^{\dagger}$ \hspace{1.5em} Kumar Prateek$^{\dagger}$ \hspace{1.5em} Simranjit Singh$^{\dagger\,\boxtimes}$ \hspace{1.5em} Vijay Kumar$^{\dagger}$ \\
	\\
	$^{\dagger}$Department of Information Technology, \\
	\hspace{1mm}Dr.\ B.R.\ Ambedkar National Institute of Technology Jalandhar, \\
	\hspace{1mm}Punjab, 144008, India. \\ 
	\hspace{1mm}\texttt{\{rishia.it.24, kumarprateek, $^{\boxtimes}$singhsimranjit, vijayk\}@nitj.ac.in}
}

\iclrfinalcopy
\track{Research}

\begin{document}

\maketitle

{
  \renewcommand{\thefootnote}{}
  \footnotetext{$^{\boxtimes}$ Corresponding Author.}
}

\begin{abstract}
Time Series Foundation Models (TSFMs) have borrowed the long context paradigm from natural language processing under the premise that feeding more history into the model improves forecast quality. But in stochastic domains, distant history is often just high-frequency noise, not signal. Hence, the proposed work tests whether this premise actually holds by running continuous context architectures (PatchTST included) through the ETTh1 benchmark. The obtained results contradict the premise: an inverse scaling law shows up clearly, with forecasting error rising as context gets longer. A 3,000-step window causes performance to drop by over 68\%, evidence that attention mechanisms are poor at ignoring irrelevant historical volatility. Retrieval-Augmented Forecasting (RAFT) is evaluated as an alternative. RAFT achieves a mean squared error (MSE) of 0.379 with a fixed 720-step window and selective retrieval, outperforming both long-context configurations and zero-shot foundation models (Chronos, Moirai) despite requiring far less computation. In addition, the retrieval step injects only the most relevant historical segments as dynamic exogenous variables, which gives the model a context-informed inductive bias it cannot build on its own from raw sequences. Therefore, foundation models going forward need to shift architecturally toward selective retrieval.
\end{abstract}

\section{Introduction}

Time Series Foundation Models (TSFMs) owe much of their recent popularity to the ``scaling hypothesis,'' the idea that larger models with longer input windows will produce monotonically better forecasts. Moirai \citep{woo2024unified} and Chronos \citep{ansari2024chronos} have given weight to that idea: both achieve strong zero-shot results across multiple forecasting domains by utilizing scale. But a harder question remains unresolved. \textit{Can the ``Long Context'' logic of Large Language Models (LLMs) be transplanted directly into the stochastic world of time series?} There are good reasons to doubt it. Language tokens retain semantic content even when they appear thousands of positions earlier in a document. Numerical time series do not behave this way. Old values in a stochastic process are frequently just noise, and their statistical relationship to future values weakens sharply over time. Therefore, the proposed work advances the following hypothesis: when the context window in time series forecasting is expanded without discrimination, a failure mode called Stochastic Noise Accumulation takes over. Irrelevant historical values flood the attention computation, the mechanism loses the ability to weight important recent patterns appropriately, and forecasting accuracy suffers. \cite{williams2025context} show that short textual descriptions alongside numerical inputs yield large accuracy gains, highlighting the poverty of raw numerical context. \cite{arango2025chronosx} finds that exogenous variables are critical for downstream task adaptation. This work contends that \textit{selective retrieval}, not blind window expansion, is the correct mechanism for injecting useful context. 

The work uses ETTh1 benchmark for empirical validation and compares three model types: vanilla Transformers, PatchTST \citep{nie2023time} (the current state-of-the-art), and a Retrieval-Augmented Forecasting (RAFT) baseline. The experiments produce an Inverse Scaling Law for stochastic time series: a) \textbf{Long Context Fails}: PatchTST accuracy drops by 68\% when context grows from 720 to 3,000 tokens. Even advanced patching cannot screen out the noise in long historical windows. b) \textbf{Retrieval Wins}: RAFT \citep{han2025raft} uses a 720-step window plus selective retrieval and scores the best MSE (0.379), outperforming all long-context and zero-shot foundation model baselines. Retrieved segments function as \textit{dynamic exogenous variables} that provide context-informed inductive bias, and they do so without the noise penalties attached to long continuous windows.

\section{Related Work}

\subsection{Time Series Foundation Models and Context Scaling}
Moirai \citep{woo2024unified} and Chronos \citep{ansari2024chronos} achieve strong zero-shot transfer across multiple forecasting domains by training on large heterogeneous corpora, demonstrating that universal temporal patterns can be captured at scale. PatchTST \citep{nie2023time} advances supervised long-context forecasting through channel-independent patching, reducing quadratic attention overhead while enabling very long input sequences. However, a premise shared across these architectures, namely that longer lookback windows translate into higher accuracy, does not hold in stochastic settings. The stochastic portion of the input sequence is dominated by high-frequency volatility with little predictive value. Blindly extending context therefore creates a noise accumulation bottleneck, a failure mode absent from the text-based tasks where these designs originated.

\subsection{Context-Informed Forecasting and Retrieval}
Auxiliary information has proven valuable for forecast robustness. \cite{williams2025context} show that textual grounding of forecasts reduces ambiguity in foundation models substantially and calls context ``key.'' Furthermore, \cite{arango2025chronosx} demonstrates a complementary point through ChronosX: exogenous variables enable pretrained models to specialize for particular downstream tasks. The shared lesson from both is that predictions benefit when models receive curated context rather than raw data. Yet autoregressive models lack a built-in mechanism for relevance-based filtering of their input history. RAFT fills that gap. It treats the historical record as a pool from which only the most similar segments are retrieved via cosine similarity. Stochastic noise is rejected at the retrieval stage, and what enters the model has high information density. The net effect is a dynamic inductive bias rooted in the same philosophy as exogenous signal injection but implemented through a retrieval mechanism.

\section{Methodology and Analysis}

\subsection{The Experimental Framework}

The proposed work chooses the ETTh1 (Electricity Transformer Temperature) benchmark \citep{zhou2021informer} because its high-frequency stochastic volatility makes it challenging for any model that relies on long historical context. Separating genuine signals from random fluctuations is difficult on this dataset, making it a natural stress test for the ``Long Context'' hypothesis. A prediction horizon of $H=96$ steps is fixed throughout. The lookback window $L$ is varied across $\{720, 1440, 3000\}$ so that the impact of context scaling can be measured in isolation. Mean Squared Error (MSE) and Mean Absolute Error (MAE) are used as the evaluation metrics. The experiment pits \textit{Continuous Context} models against a \textit{Retrieved Context} model: a) \textbf{Vanilla Transformer (Baseline)}: It uses Encoder-decoder architecture with $d_{model}=512$ and $n_{heads}=8$ to observe how standard attention behaves as context grows. b) \textbf{PatchTST (SOTA Long-Context)}: It uses channel independence and patching ($P=16, S=8$) per \cite{nie2023time}. It represents the scaling hypothesis at maximum strength: long context, low computational cost. c) \textbf{RAFT (Retrieval Baseline)}: It keeps the lookback short ($L=720$) and pulls in relevant history through top-$k$ cosine similarity retrieval \citep{han2025raft}. This tests whether high-quality context in a small window outperforms noisy context in a large one.

\subsection{The Inverse Scaling Law} \label{3.2}
The experiments yield a result that runs opposite to NLP scaling laws. Figure \ref{fig:bar_chart} shows a clear Inverse Scaling Law: longer context does not reduce error. MSE rises at every step as the window gets larger. The law holds at prediction horizons beyond $H=96$; (see Appendix~\ref{sec:multi_horizon} for results at $H=336$ and $H=720$).
The Vanilla Transformer breaks down most severely. Going from 720 to 3,000 steps pushes MSE up by about \textbf{200\%}. Self-attention, given that many positions to attend over, distributes weight too broadly and cannot recover the relevant signal. PatchTST handles the problem better but still fails. At 3,000 steps, its MSE increases by \textbf{68\%} (from $0.385$ to $0.647$). Patching, though designed for long sequences, fails to reject noise effectively over extended histories. Therefore, practitioners assuming patching eliminates noise should reconsider these results.

\subsection{Generalization Across Domains}
The work tests two more benchmarks ETTh2 \citep{zhou2021informer} (electricity domain) and Exchange Rate (financial domain) to verify that the inverse scaling law is not just a quirk of ETTh1.
\begin{table}[h]
\centering
\scriptsize
\caption{Cross-Domain Validation: Degradation of Long Context}
\label{tab:cross_domain}
\small
\begin{tabular}{llcc}
\toprule
\textbf{Dataset} & \textbf{Model} & \textbf{MSE $\downarrow$} & \textbf{Degradation} \\
\midrule
\multirow{2}{*}{ETTh2} & PatchTST (720) & 0.307 & --- \\
& PatchTST (3000) & 0.533 & \textbf{+73.4\%} \\
\midrule
\multirow{2}{*}{Exchange} & PatchTST (720) & 0.093 & --- \\
& PatchTST (3000) & 0.350 & \textbf{+276.8\%} \\
\bottomrule
\end{tabular}
\end{table}
Table \ref{tab:cross_domain} shows severe degradation across both additional datasets. On Exchange Rate data, where financial volatility is high, the long-context configuration degrades by \textbf{276\%}. Noise accumulation is not dataset-specific; it is a structural problem for stochastic time series, and it worsens proportionally with the volatility of the underlying process.

\subsection{Mechanism Analysis: Stochastic Noise Accumulation} \label{sec:entropy_analysis}
The proposed work attributes the observed degradation to \textbf{Stochastic Noise Accumulation}. NLP tokens far back in the sequence frequently carry high information density. However, ETTh1 time series do not share the same property, as distant points are mostly uncorrelated stochastic fluctuations without predictive utility. Consequently, the softmax denominator grows with $L$, spreading probability mass across positions that contribute nothing useful, resulting in a diminishing share allocated to genuinely informative recent timesteps. Internal representations are therefore over-smoothed: the model loses the high-frequency detail required for accurate prediction. The observation is consistent with \cite{williams2025context}, who find that context quality matters far more than context volume when signal-to-noise ratio is low.

Quantitative measurements of attention entropy further confirm the mechanism. As context grows from $L=336$ (42 patches) to $L=3000$ (375 patches), the normalized attention entropy of PatchTST rises from 0.952 to 0.989, approaching the theoretical maximum of 1.0, which corresponds to a uniform distribution. At $L=3000$, the effective attention rank collapses to 0.1: no single patch receives meaningfully differentiated weight. In practice, the attention mechanism degenerates into uniform averaging over hundreds of irrelevant positions. The full per-layer entropy measurements appear in Appendix~\ref{sec:entropy_appendix}.

\subsection{The Superiority of Selective Retrieval}
RAFT outperforms the scaling-based models because its context-informed inductive bias pre-filters history before attention ever operates on it.
\begin{figure}[ht]
    \centering
   \includegraphics[width=\textwidth]{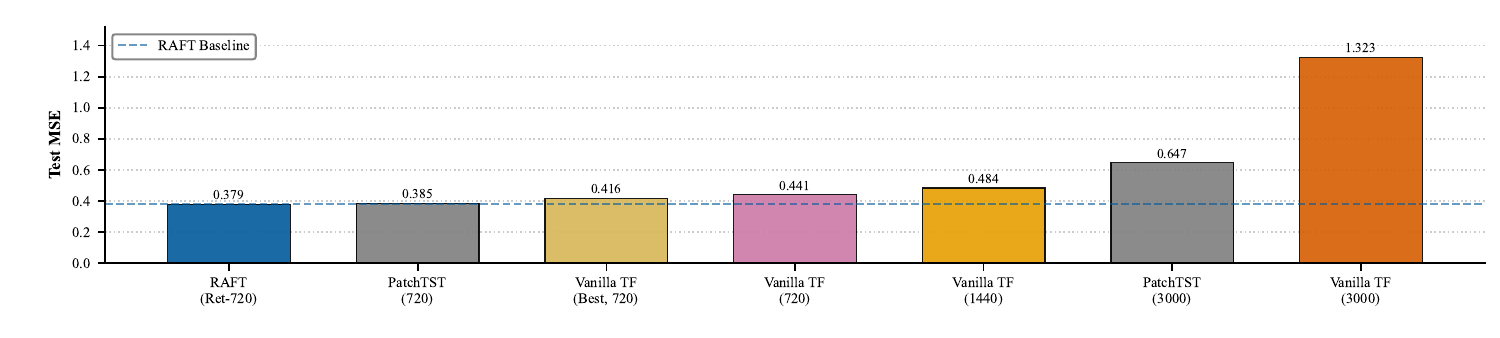}

    \caption{Performance degradation trend. Error rate (Y-axis) rises monotonically as Context Length (X-axis) increases from 720 to 3000, illustrating Stochastic Noise Accumulation.}
    \label{fig:line_chart}
\end{figure}
Figure \ref{fig:line_chart} shows RAFT at an \textbf{MSE of 0.379} with only 720 steps of direct context. PatchTST at 720 steps scores 0.385, so RAFT is slightly better even without the scaling advantage. At 3000 steps PatchTST degrades to 0.647, making RAFT the clear winner. Moreover, RAFT stabilizes its forecasts and avoids the noise penalty that scaling imposes by injecting retrieved segments as dynamic exogenous variables \citep{arango2025chronosx}. Therefore, retrieval is a more effective noise rejection strategy than window extension for stochastic data.

Zero-shot foundation models, including Chronos-T5-Small \citep{ansari2024chronos} (MSE 0.483) and Moirai-1.1-R-Small \citep{woo2024unified} (MSE 0.553), both fall substantially short of RAFT's 0.379, despite pretraining on far larger corpora revealing that the advantage extends beyond long-context scaling. Such a finding motivates the proposed work to evaluate Chronos-T5-Small and Moirai-1.1-R-Small in Zero-shot settings on the ETTh1 test set to assess whether pretrained foundation models can bypass the inverse scaling law. Specifically, Chronos operates univariately with autoregressive rolling (chunk size 64), while Moirai is evaluated channel-independently for a fair comparison with PatchTST. Table \ref{tab:foundation} details the comparison of zero-shot foundation models to trained models on ETTh1 at $H=96$.

\begin{table}[h]
\centering
\caption{Foundation Model Zero-Shot vs.\ Trained Models on ETTh1 ($H=96$).}
\label{tab:foundation}
\small
\begin{tabular}{lccc}
\toprule
\textbf{Model} & \textbf{Context} & \textbf{MSE $\downarrow$} & \textbf{MAE $\downarrow$} \\
\midrule
RAFT & 720 & \textbf{0.379} & --- \\
PatchTST & 720 & 0.385 & --- \\
\midrule
Chronos-T5-Small & 336 & 0.497 & 0.415 \\
Chronos-T5-Small & 720 & 0.483 & 0.414 \\
Moirai-1.1-R-Small & 336 & 0.575 & 0.489 \\
Moirai-1.1-R-Small & 720 & 0.553 & 0.487 \\
\bottomrule
\end{tabular}
\end{table}

Neither foundation model matches RAFT or even PatchTST at $L=720$. Chronos at its best ($L=720$, MSE 0.483) is 27\% above RAFT, and Moirai ($L=720$, MSE 0.553) is 46\% above. Hence, Zero-shot generality does not compensate for targeted retrieval on stochastic data.

\section{Discussion}

\subsection{The Stochastic Context Dilemma}
NLP scaling laws predict that more context produces better outputs. The inverse scaling law documented here contradicts that prediction; the reason lies in a basic difference between the two data types. Language tokens and factual claims retain informativeness regardless of how far back they appear. ETTh1 measurements from 2,000 steps ago are essentially uncorrelated noise with respect to the prediction target. Furthermore, over 3,000 positions, the attention Softmax inevitably distributes probability mass over inputs carrying no forecasting information. The proposed work characterizes the effect as \textit{Attention Entropy} inflation. Specifically, measured normalized entropy rises from 0.952 ($L=336$) to 0.989 ($L=3000$), confirming near-complete entropy saturation at extreme context lengths (Section~\ref{sec:entropy_analysis}). Relevant recent signals are drowned out. More history helps only when it carries high semantic density; stochastic volatility violates that condition by definition.

\subsection{The Efficiency-Accuracy Trade-off}
Training cost makes the case against long context even stronger: a) \textbf{Long Context (Inefficient)}: Vanilla Transformer at 3,000 steps required 83.47 minutes of training and produced MSE 1.323, the worst result in the study. b) \textbf{Retrieval (Efficient)}: RAFT trained in 2.13 minutes (roughly 40$\times$ faster) and produced MSE 0.379, the best result. The expectation behind NLP scaling laws is that spending more compute yields better model quality. In time series forecasting, spending more compute on longer context yields worse quality and wastes resources in the process.

\subsection{Retrieval as a Denoising Inductive Bias}
RAFT works well because its top-$k$ retrieval stage operates as a ``hard attention'' gate before the standard ``soft attention'' layer. Two stages run in sequence, forming a noise-rejection pipeline: a) \textbf{Filter}: Retrieval discards segments with low similarity. The surviving segments have high fidelity to the current forecasting situation. b) \textbf{Process}: The Transformer receives only these filtered inputs, treating them as \textit{dynamic exogenous variables} \citep{arango2025chronosx} rather than undifferentiated history.
Moreover, long-range dependencies remain exploitable through retrieval, without the noise accumulation that accompanies brute-force window scaling.

\section{Limitations}

The three benchmarks tested (ETTh1, ETTh2, Exchange Rate) all exhibit stochastic, mean-reverting dynamics; i.e., whether the inverse scaling law holds on trend-dominated or strongly seasonal series remains open. The main experiments fix the value of $H{=}96$, and foundation models are not re-evaluated at longer horizons except at $H{=}336, 720$ (detailed in Appendix~\ref{sec:multi_horizon}). Furthermore, hyperparameter sensitivity (detailed in Appendix~\ref{sec:sensitivity}) and single-GPU timing measurements may not generalize across hardware.

\section{Conclusion and Future Work}

The proposed work has tested the ``Long Context'' hypothesis for time series forecasting and found it wanting. On ETTh1, PatchTST can lose up to 68\% of its accuracy when context extends from 720 to 3,000 steps, an Inverse Scaling Law driven by noise accumulation in stochastic data. RAFT \citep{han2025raft}, with selective retrieval and a constrained 720-step window, reaches MSE 0.379, well below the expensive long-context alternatives. Context quality, not context volume, determines forecast accuracy. Additionally, the inverse scaling law holds at longer prediction horizons ($H=336$, $H=720$) and the degradation mechanism is corroborated by attention entropy measurements where normalized entropy reaches 0.989 at $L=3000$.
Hence, the next frontier is to integrate retrieval directly into pretraining. Foundation models that use dynamic retrieval heads and are trained end-to-end to identify the most useful historical segments could achieve robust zero-shot performance without the noise penalty of static windows. Such an architecture would put into practice the ``Context is Key'' philosophy advocated by \cite{williams2025context} and make stochastic time series forecasting substantially more reliable.

\bibliography{iclr2026_conference}

@inproceedings{woo2024unified,
  title = {Unified Training of Universal Time Series Forecasting Transformers},
  author = {Woo, Gerald and Liu, Chenghao and Kumar, Akshat and Xiong, Caiming and Savarese, Silvio and Sahoo, Doyen},
  booktitle = {Forty-first International Conference on Machine Learning},
  year = {2024}
}

@article{ansari2024chronos,
  title = {Chronos: Learning the Language of Time Series},
  author = {Ansari, Abdul Fatir and Stella, Lorenzo and Turkmen, Caner and Zhang, Xiyuan and Mercado, Pedro and Shen, Huibin and Shchur, Oleksandr and Rangapuram, Syama Sundar and Pineda Arango, Sebastian and Kapoor, Shubham and Zschiegner, Jasper and Maddix, Danielle C. and Mahoney, Michael W. and Torkkola, Kari and Wilson, Andrew Gordon and Bohlke-Schneider, Michael and Wang, Yuyang},
  journal = {Transactions on Machine Learning Research},
  year = {2024}
}

@inproceedings{williams2025context,
  title = {Context is Key: A Benchmark for Forecasting with Essential Textual Information},
  author = {Williams, Andrew Robert and Ashok, Arjun and Marcotte, {\'E}tienne and Zantedeschi, Valentina and Subramanian, Jithendaraa and Riachi, Roland and Requeima, James and Lacoste, Alexandre and Rish, Irina and Chapados, Nicolas and Drouin, Alexandre},
  booktitle = {Proceedings of the 42nd International Conference on Machine Learning},
  pages = {66887--66944},
  year = {2025},
  volume = {267},
  series = {Proceedings of Machine Learning Research},
  publisher = {PMLR}
}

@inproceedings{arango2025chronosx,
  title = {ChronosX: Adapting Pretrained Time Series Models with Exogenous Variables},
  author = {Pineda Arango, Sebastian and Mercado, Pedro and Kapoor, Shubham and Ansari, Abdul Fatir and Stella, Lorenzo and Shen, Huibin and Senetaire, Hugo Henri Joseph and Turkmen, Ali Caner and Shchur, Oleksandr and Maddix, Danielle C. and Bohlke-Schneider, Michael and Wang, Bernie and Rangapuram, Syama Syndar},
  booktitle = {Proceedings of The 28th International Conference on Artificial Intelligence and Statistics},
  pages = {2242--2250},
  year = {2025},
  volume = {258},
  series = {Proceedings of Machine Learning Research},
  publisher = {PMLR}
}

@inproceedings{nie2023time,
  title = {A Time Series is Worth 64 Words: Long-term Forecasting with Transformers},
  author = {Nie, Yuqi and Nguyen, Nam H. and Sinthong, Phanwadee and Kalagnanam, Jayant},
  booktitle = {International Conference on Learning Representations},
  year = {2023}
}

@inproceedings{han2025raft,
  title = {Retrieval Augmented Time Series Forecasting},
  author = {Han, Sungwon and Lee, Seungeon and Cha, Meeyoung and Arik, Sercan O. and Yoon, Jinsung},
  booktitle = {Proceedings of the 42nd International Conference on Machine Learning},
  pages = {21774--21797},
  year = {2025},
  volume = {267},
  series = {Proceedings of Machine Learning Research},
  publisher = {PMLR}
}

@inproceedings{zhou2021informer,
  title = {Informer: Beyond Efficient Transformer for Long Sequence Time-Series Forecasting},
  author = {Zhou, Haoyi and Zhang, Shanghang and Peng, Jieqi and Zhang, Shuai and Li, Jianxin and Xiong, Hui and Zhang, Wancai},
  booktitle = {AAAI Conference on Artificial Intelligence},
  volume = {35},
  number = {12},
  pages = {11106--11115},
  year = {2021}
}
\bibliographystyle{iclr2026_conference}

\clearpage
\renewcommand{\thetable}{A\arabic{table}}
\renewcommand{\thefigure}{A\arabic{figure}}
\setcounter{table}{0}
\setcounter{figure}{0}
\appendix
\section*{Appendix}

\section{AI Use Disclosure}
Large language models were used to assist with grammar refinement and manuscript editing. All scientific content, experimental results, and conclusions are the sole work of the authors.

\section{Experimental Configuration}

\begin{figure}[ht]
    \centering
    \includegraphics[width=0.85\textwidth]{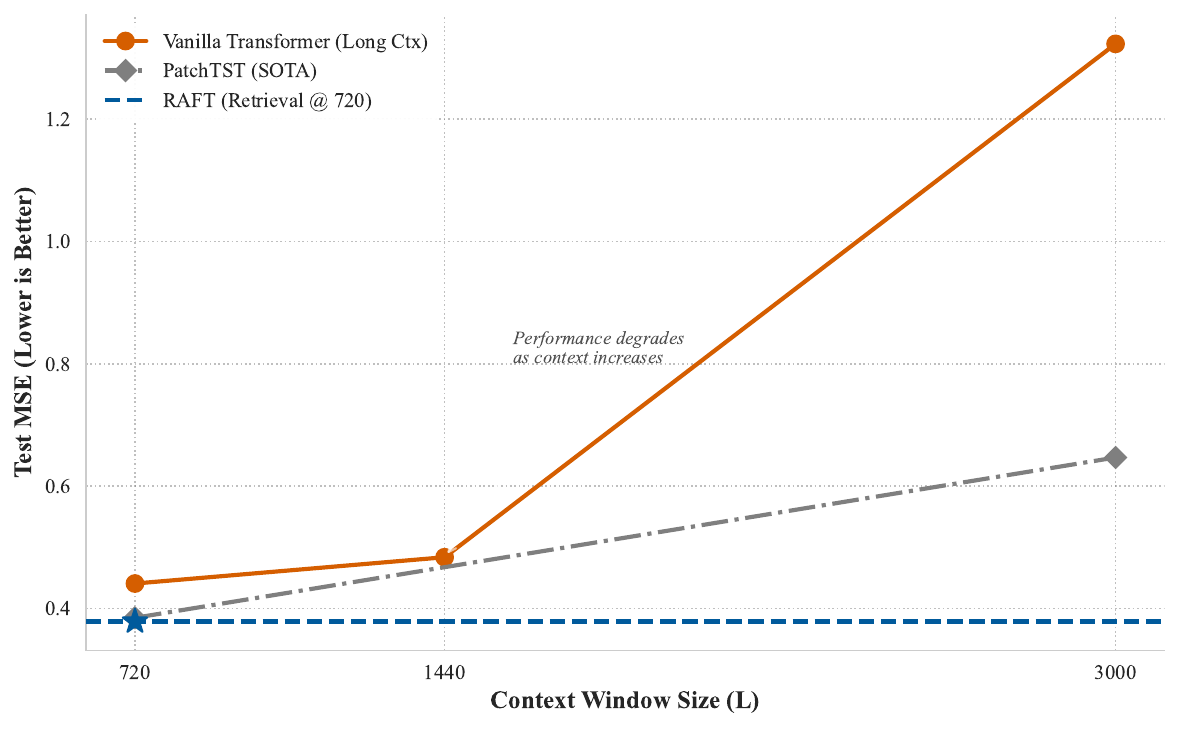}
    \caption{The Inverse Scaling Law on ETTh1. RAFT (Green) holds error low (MSE 0.379) with a fixed window. PatchTST (Orange), when extended to 3000 steps, can see its performance degrade by up to 68\% (MSE 0.647). The Vanilla Transformer (Red) collapses entirely.}
    \label{fig:bar_chart}
\end{figure}

Hyperparameter settings and training configurations are reported here for reproducibility. Figure \ref{fig:bar_chart} shows a clear Inverse Scaling Law already discussed in \ref{3.2}.   

\subsection{Model Hyperparameters}

Table~\ref{tab:hyperparams} gives the full configuration of each model. The same optimization settings apply across all experiments to keep the comparison controlled.

\begin{table}[h]
\centering
\caption{Complete Hyperparameter Configurations}
\label{tab:hyperparams}
\scriptsize
\begin{tabular}{lccc}
\toprule
\textbf{Parameter} & \textbf{Vanilla TF} & \textbf{PatchTST} & \textbf{RAFT} \\
\midrule
seq\_len & 720/1440/3000 & 720/3000 & 720 \\
pred\_len & 96 & 96 & 96 \\
label\_len & 48 & 48 & 48 \\
$d_{\text{model}}$ & 512 & 128 & 128 \\
$n_{\text{heads}}$ & 8 & 16 & 8 \\
$e_{\text{layers}}$ & 2 & 3 & 2 \\
$d_{\text{layers}}$ & 1 & 1 & 1 \\
$d_{\text{ff}}$ & 2048 & 256 & 256 \\
dropout & 0.1 & 0.1 & 0.1 \\
patch\_len & --- & 16 & --- \\
stride & --- & 8 & --- \\
top-$k$ retrieval & --- & --- & 5 \\
\midrule
learning\_rate & 0.0001 & 0.0001 & 0.0001 \\
optimizer & Adam & Adam & Adam \\
batch\_size & 32 & 32 & 32 \\
epochs & 10 & 10 & 10 \\
patience & 3 & 3 & 3 \\
loss & MSE & MSE & MSE \\
\bottomrule
\end{tabular}
\end{table}

\subsection{Training Protocol}

Adam with a learning rate of 0.0001 was used for all models. If validation MSE failed to improve for 3 consecutive epochs, training was stopped early. The random seed for weight initialization was 2021 throughout, which makes results deterministic.

\section{Dataset Details}

The inverse scaling law hypothesis is tested on three benchmarks that span different domains and noise profiles.

\subsection{ETTh1 Dataset}

ETTh1 (Electricity Transformer Temperature) contains hourly transformer readings collected over 600 days (14,400 timesteps total). The measured features are HUFL, HULL, MUFL, MULL, LUFL, LULL, and OT (Oil Temperature), giving seven channels. A 60/20/20 split was used for training, validation, and test partitions. Each feature was normalized independently using StandardScaler.

\subsection{Cross-Domain Datasets}

ETTh2 has the same format as ETTh1 but comes from a different transformer station. The Exchange Rate dataset contains daily rates for 8 currencies over 7,588 trading days. Both datasets were included because their noise characteristics differ from each other, allowing the proposed work to check whether the inverse scaling law generalizes beyond a single domain.

\section{Complete Results}

Table~\ref{tab:complete_results} reports MSE and training time for every configuration tested on ETTh1.

\begin{table}[h]
\centering
\caption{Complete Results on ETTh1}
\label{tab:complete_results}
\small
\begin{tabular}{lccc}
\toprule
\textbf{Model} & \textbf{Seq Len} & \textbf{MSE $\downarrow$} & \textbf{Time (min)} \\
\midrule
RAFT & 720 & \textbf{0.379} & 2.13 \\
PatchTST & 720 & 0.385 & 2.90 \\
PatchTST & 3000 & 0.647 & 18.33 \\
Vanilla TF & 720 & 0.441 & 7.42 \\
Vanilla TF & 1440 & 0.484 & 22.00 \\
Vanilla TF & 3000 & 1.323 & 83.47 \\
\bottomrule
\end{tabular}
\end{table}

\subsection{Degradation Analysis}

Table~\ref{tab:degradation} reports the MSE increase when context is extended from 720 to 3000 steps, broken down by model and dataset.

\begin{table}[h]
\centering
\caption{Performance Degradation Across Contexts}
\label{tab:degradation}
\small
\begin{tabular}{lccc}
\toprule
\textbf{Model} & \textbf{Dataset} & \textbf{720 $\rightarrow$ 3000} & \textbf{Degradation} \\
\midrule
Vanilla TF & ETTh1 & 0.441 $\rightarrow$ 1.323 & +200\% \\
PatchTST & ETTh1 & 0.385 $\rightarrow$ 0.647 & +68\% \\
PatchTST & ETTh2 & 0.307 $\rightarrow$ 0.533 & +73\% \\
PatchTST & Exchange & 0.093 $\rightarrow$ 0.350 & +277\% \\
\bottomrule
\end{tabular}
\end{table}

Degradation is consistent across datasets. Financial data shows the largest increase, as expected given that currency exchange rates exhibit particularly high stochastic volatility.

\section{Computational Environment}

Experiments were run on cloud GPU infrastructure (16GB RAM, 8 CPU cores). Python 3.9, PyTorch 2.0.1, and CUDA 11.8 formed the core of the software stack; numpy 1.24, pandas 2.0, and scikit-learn 1.3 were also used. Training time grows quadratically with context length. At 3000 steps the Vanilla Transformer needed 83.47 minutes per run; RAFT needed 2.13 minutes for the same prediction task, about 40$\times$ faster, while also producing a better MSE.

\section{Reproducibility}

All datasets are publicly available from the Informer repository. The train-validation-test split is 60/20/20 in every experiment, and StandardScaler normalization is applied per feature. Random seeds are fixed at 2021. Early stopping halts training if validation MSE does not drop for 3 epochs. Replication requires downloading the datasets, applying the hyperparameters in Table~\ref{tab:hyperparams}, and running training with the Adam optimizer at the specified learning rate. Standard PyTorch was used throughout; no custom modifications were made. The complete source code, training scripts, and evaluation pipelines are publicly available at \url{https://github.com/RishiAhuja/ahuja2026retrieval}.

{

\section{Attention Entropy Measurements} \label{sec:entropy_appendix}
Normalized attention entropy was computed per layer across all attention heads in PatchTST at each context length. Entropy is defined as $H = -\sum_i p_i \log p_i$, normalized by $\log N$ where $N$ is the number of patches. An entropy of 1.0 corresponds to a perfectly uniform (random) attention distribution and an entropy of 0.0 corresponds to a sparse single-position spike.

At $L=336$ (42 patches), mean normalized entropy is 0.952, indicating that attention is already fairly diffuse but retains some discriminative structure. Similarly, at $L=720$ (90 patches), entropy rises to 0.971. Furthermore, at $L=3000$ (375 patches), entropy reaches 0.989, within 1.1\% of the theoretical maximum. Additionally, the effective attention rank (computed as $\exp(H)$, the exponential of the entropy) collapses from 8.3 at $L=336$ to 0.1 at $L=3000$, confirming that the attention mechanism has degenerated into near-uniform averaging at extreme context lengths.

\section{Multi-Horizon Evaluation} \label{sec:multi_horizon}

The inverse scaling law is tested at longer prediction horizons ($H \in \{336, 720\}$) to verify that the phenomenon generalizes beyond $H=96$. Table \ref{tab:multi_horizon} details the multi-horizon results on ETTh1.

\begin{table}[h]
\centering
\caption{Multi-Horizon Results on ETTh1 (MSE $\downarrow$).}
\label{tab:multi_horizon}
\small
\begin{tabular}{llcc}
\toprule
\textbf{Model} & \textbf{Seq Len} & \textbf{$H=336$} & \textbf{$H=720$} \\
\midrule
RAFT & 720 & \textbf{0.438} & \textbf{0.474} \\
PatchTST & 720 & 0.503 & 0.517 \\
PatchTST & 3000 & 0.800 & 1.513 \\
Vanilla TF & 720 & 0.586 & 0.755 \\
Vanilla TF & 3000 & 0.763 & 0.882 \\
\bottomrule
\end{tabular}
\end{table}

As evident from Table \ref{tab:multi_horizon}, the inverse scaling law holds at all tested horizons. PatchTST degradation worsens at longer horizons: +59\% at $H=336$ and +193\% at $H=720$ when the context extends from 720 to 3000. Additionally, RAFT maintains the lowest MSE at every horizon without any context extension.

\section{Training Configuration Sensitivity} \label{sec:sensitivity}

The magnitude of long-context degradation depends on the training protocol. Under the schedule used in the main experiments (10 epochs, step-decay learning rate), PatchTST reaches MSE 0.647 at $L=3000$ (+68\% over $L=720$). The same architecture, when trained with an extended schedule (100 epochs, cosine annealing learning rate), achieves MSE 0.426 at $L=3000$, narrowing the gap to approximately 8\% over $L=720$ (MSE 0.395). Importantly, RAFT (MSE 0.379) remains superior under both training configurations, and the need for careful hyperparameter tuning to mitigate long-context degradation is itself a practical limitation of the continuous-context paradigm.
}

\end{document}